\documentclass[runningheads]{llncs}
\usepackage[T1]{fontenc}
\usepackage{graphicx}
\usepackage{todonotes}
\usepackage{booktabs}
\usepackage{hyperref}
\usepackage{color}

\begin{document}
\title{BioGraphletQA: Knowledge-Anchored Generation of Complex QA Datasets}
\author{Richard A. A. Jonker \inst{1}\orcidID{0000-0002-3806-6940} \and
Bárbara Maria Ribeiro de Abreu Martins \inst{2}\orcidID{0009-0005-5183-2469} \and
Sérgio Matos \inst{1}\orcidID{0000-0003-1941-3983}}

\authorrunning{Jonker et al.}
\institute{
IEETA, DETI, LASI, University of Aveiro, Aveiro, Portugal \\
\email{\{richard.jonker, aleixomatos\}@ua.pt}
\and
USF Atlântico Norte, Portugal\\
\email{bmmartins@ulsra.min-saude.pt}
}
\maketitle              % typeset the header of the contribution
\begin{abstract}

This paper presents a principled and scalable framework for systematically generating complex Question Answering (QA) data. In the core of this framework is a graphlet-anchored generation process, where small subgraphs from a Knowledge Graph (KG) are used in a structured prompt to control the complexity and ensure the factual grounding of questions generated by Large Language Models. The first instantiation of this framework is BioGraphletQA, a new biomedical KGQA dataset of 119,856 QA pairs. Each entry is grounded in a graphlet of up to five nodes from the OREGANO KG, with most of the pairs being enriched with relevant document snippets from PubMed. We start by demonstrating the framework's value and the dataset's quality through evaluation by a domain expert on 106 QA pairs, confirming the high scientific validity and complexity of the generated data. Secondly, we establish its practical utility by showing that augmenting downstream benchmarks with our data improves accuracy on PubMedQA from 49.2\% to 68.5\% in a low-resource setting, and on MedQA from a 41.4\% baseline to 44.8\% in a full-resource setting. Our framework provides a robust and generalizable solution for creating critical resources to advance complex QA tasks, including MCQA and KGQA. All resources supporting this work, including the dataset (\url{https://zenodo.org/records/17381119}) and framework code (\url{https://github.com/ieeta-pt/BioGraphletQA}), are publicly available to facilitate use, reproducibility and extension.

\keywords{Knowledge Graph \and
Biomedical Question Answering \and
Synthetic Data \and
Large Language Models \and
Information Retrieval.}
\end{abstract}

\section{Introduction}

Question answering (QA) systems have benefited immensely from advances in large language models (LLMs), particularly Transformer-based architectures \cite{vaswani2017attention}. However, despite their success, LLMs struggle with factual consistency, often generating hallucinated or inaccurate responses \cite{10.1145/3571730,10.1145/3703155}. One promising approach to mitigate these issues is the use of Knowledge Graph Question Answering (KGQA) datasets. Traditional KGQA datasets, however, are either manually curated—making them costly and time-intensive \cite{gu2021beyond}—or template-based, which often limits their diversity and generalizability \cite{banerjee2023dblp}.

In the biomedical domain, the problem of hallucinations can lead to dangerous outcomes such as misdiagnoses, unsafe treatment recommendations, and compromised patient safety. Although several biomedical KGs exist—such as OREGANO KG \cite{boudin2023oregano}, CKG \cite{santos2022knowledge}, MonarchKG \cite{10.1093/nar/gkad1082}, and PrimeKG \cite{chandak2022building}—most KGQA research has focused on large open-domain KGs like Freebase \cite{bollacker2008freebase} and Wikidata \cite{vrandevcic2014wikidata}, which often lack the granularity and reliability required for biomedical decision-making. To date, only one large-scale synthetic biomedical KGQA dataset has been developed \cite{yan2024bridging}, generated using an LLM with graphlets from PrimeKG, underscoring the need for more robust, domain-specific QA resources.

% \begin{figure}[ht]

To address this limitation, we propose a framework for the systematic generation of KGQA data. The framework is centered on a graphlet-anchored generation process, wherein small, coherent subgraphs (graphlets) are systematically extracted from a KG. These graphlets serve as support to guide an LLM in formulating intricate questions, ensuring the factual grounding of the generated output by constraining it to the relations within the subgraph, while simultaneously leveraging the linguistic capabilities of the LLM to achieve a high degree of complexity and naturalness, unachievable through conventional template-based methods.

The first instantiation of this framework is BioGraphletQA, a new biomedical KGQA dataset comprising 119,856 QA pairs. Each entry is explicitly grounded in a multi-node graphlet (3-5 nodes) derived from the OREGANO KG (v2.1) and is subsequently enriched with relevant document snippets from PubMed. The quality and utility of the dataset were validated through a dual evaluation. First, a qualitative assessment by a domain expert confirmed the high scientific validity and complexity of a small sample of the generated data. Second, its quantitative utility was established by using BioGraphletQA as an augmentation resource for downstream tasks. This augmentation resulted in performance gains, improving mean accuracy on the PubMedQA benchmark from 49.2\% to 68.5\% in a low-resource setting, and increasing the baseline performance on MedQA from 41.4\% to 44.8\%.

Our primary contributions are twofold: 
\begin{enumerate}
    \item We present a robust and generalizable data generation framework that is adaptable to other KGs and domains.
    \item We present BioGraphletQA, a large-scale, complex KGQA dataset intended to support future research in the biomedical domain, with use cases in model training, as well as domain and task specific fine-tuning.

\end{enumerate}

All associated resources, including the dataset and framework code, are made publicly available to facilitate future research in complex question answering.

\section{Related Work}

Recent advances in LLMs have spurred a surge in using synthetic data generation to overcome data scarcity and privacy challenges in IR and QA tasks. For instance, Braga et al. \cite{braga2024synthetic} propose a framework that generates synthetic answers tailored for personalized community QA, demonstrating that fine-tuning on this generated data can yield performance comparable to models trained on human-curated datasets. Similarly, Tang et al. \cite{tang2023does} explores leveraging ChatGPT to generate synthetic clinical documents, reporting substantial improvements in downstream tasks like named entity recognition and relation extraction. In addition, GeMQuAD, introduced by Namboori et al. \cite{namboori2024gemquad}, employs few-shot learning with LLMs to create multilingual QA datasets, thereby enhancing performance in low-resource settings. Complementing these efforts, Wu et al. \cite{wu2024synthetic} present a synthetic multimodal question generation approach that combines the strengths of LLMs and multimodal models to produce high-quality QA pairs from diverse document types.

KGQA datasets have evolved significantly, with several notable benchmarks such as LC-QuAD \cite{dubey2019lc} and ComplexQuestions \cite{bao-etal-2016-constraint}. GrailQA \cite{gu2021beyond} and GrailQA++ \cite{dutt2023grailqa++} advanced the field by introducing a dataset specifically designed to evaluate generalization in KGQA systems across different levels of compositional complexity. Jiang et al. \cite{jiang2022knowledge} provided a comprehensive survey of KGQA methods and datasets, highlighting the challenges and opportunities in this domain.

Recent advances in biomedical KGQA include PrimeKGQA~\cite{yan2024bridging}, which contains approximately 84,000 QA pairs generated through few-shot prompting using graphlets extracted from PrimeKG. This approach builds on graphlet-based methodologies similar to those in GrailQA++~\cite{dutt2023grailqa++}. Our work follows a similar graphlet-based idea but extends it by introducing an in-depth modular prompt building strategy and an additional QA filtering phase to improve quality.

Similarly, ConvKGYarn~\cite{pradeep-etal-2024-convkgyarn} generates synthetic QA pairs by combining KG facts with slot-filled question templates. While this enables large-scale QA generation, the reliance on predefined templates can limit question diversity and contextual depth. In contrast, our dynamic node selection strategy allows the LLM to flexibly identify relevant nodes and relations within each graphlet, leading to more varied and contextually nuanced QA generation.

\section{Framework Overview}

\begin{figure*}[htbp]
    \centering
    \includegraphics[width=\linewidth]{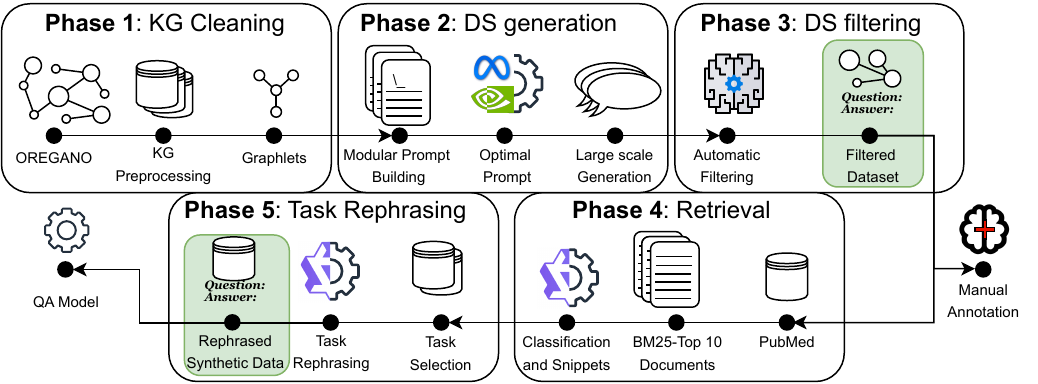}

    \caption{An overview of our framework, composed of five phases: 1) Initial cleaning of the OREGANO KG, including KG pre-processing and graphlet extraction; 2) Generation of the initial KGQA dataset, starting with a modular prompt building stage; 3) Automatic filtering stage using LLM-as-a-judge; 4) Supporting document retrieval from PubMed with BM25 using an LLM to classify the document relevance and extract snippets; 5) Task specific rephrasing for downstream tasks.}
    \label{fig:overview}
\end{figure*}

To address the challenge of creating high-quality, domain-specific QA data at scale, we propose a new multi-stage framework that relies on a modular prompt enriched with structured knowledge in the form of graphlets to provide a factual anchor to control the output of an LLM. This ensures that the generated questions are not only linguistically complex but also verifiably grounded in the underlying knowledge base.

Figure~\ref{fig:overview} provides a comprehensive overview of the framework's five sequential stages. The process begins with the preparation of a KG and the extraction of graphlets (Section~\ref{sec:kg}). These graphlets then serve as the input for the generation of the QA pairs (Section~\ref{sec:ds_gen}). To ensure high quality, the generated data undergoes automated filtering and validation process (Section~\ref{sec:filtering}). The framework concludes with stages for enriching the data with retrieved textual evidence (Section~\ref{sec:retrieval}) and aligning it for specific downstream tasks (Section~\ref{sec:rephrasing}), demonstrating its versatility.

\subsection{Knowledge Graph}
\label{sec:kg}

Developing a robust KGQA dataset requires an appropriate KG. For the initial test of the framework, the biomedical domain was chosen, as it offers unique challenges and opportunities. From the numerous existing biomedical KGs \cite{haas2024survey}, the selection criteria required a resource that balanced size and complexity while ensuring diverse node classes linked to reputable biomedical databases for comprehensive question generation. Based on these criteria, the OREGANO KG (v2.1) \cite{boudin2023oregano} was selected. It contains 88,937 nodes spanning 11 types\footnote{Note that there is also a `code' entity class, which was not utilized.} and 824,231 edges with 19 edge types.

\subsubsection{Pre-processing}

The preparation of the KG for graphlet extraction involves a two-stage pre-processing pipeline consisting of entity hydration followed by structural reduction. This stage of the pipeline is KG-specific and needs to be adapted according to the selected KG and domain.

The initial stage, hydration, is necessary to enrich the graph by resolving and updating node identifiers to their canonical names. While the OREGANO dataset includes many entity names, a significant portion required external look-up. Between December 3 and 19, 2024, various knowledge bases were systematically queried to retrieve the most current names for entities. The licenses of all source knowledge bases were also verified to permit the use and publication of these names. This process resulted in a graph with 85,655 denormalized nodes, of which 81,240 (94.85\%) were unique. Only two entities—one disease and one pathway—could not be successfully hydrated.

\begin{figure}[htbp]
    \centering
    \includegraphics[width=0.7\linewidth]{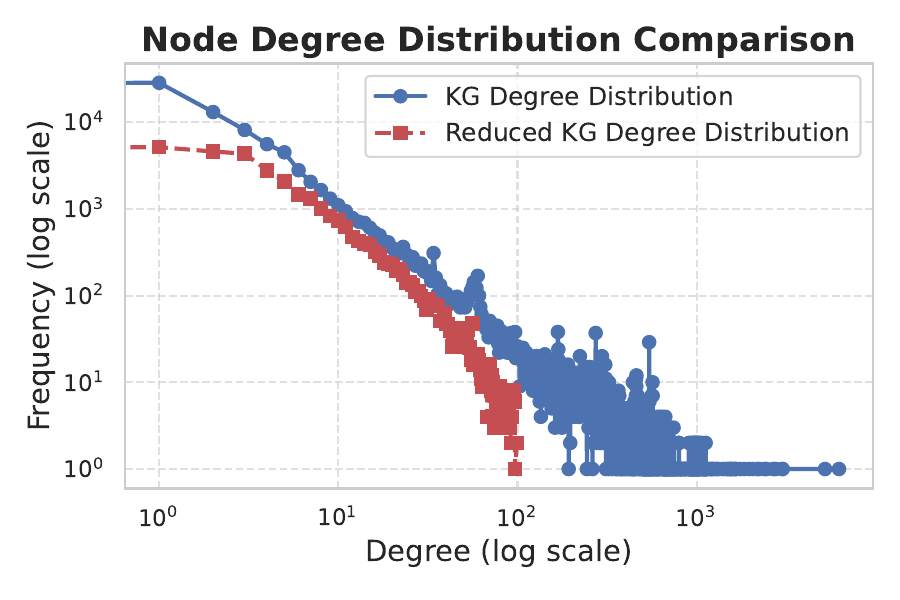}
    
    \caption{Node degree distribution of the OREGANO KG and the reduced version.}
    \label{fig:Deg_distribution}
\end{figure}

Following hydration, the reduction stage serves to refine the graph structure and improve the efficiency of subsequent processing steps. An analysis of the node degree distribution revealed a large number of nodes with a degree of one (``edge nodes'') and a small subset with very high degrees (``hub nodes''), as shown in Figure~\ref{fig:Deg_distribution}. Edge nodes were hypothesized to offer limited utility for generating complex questions due to their sparse connectivity, while hub nodes risk introducing redundancy by appearing in an excessive number of graphlets. To mitigate these issues, the graph was filtered by removing all nodes with a degree greater than 100 or less than 3. This reduction step was designed to enhance the variability of nodes in the dataset and preserve meaningful structural complexity while making the graph more computationally tractable. The final, processed graph comprises 41,115 nodes and 129,992 edges. The updated node degree is illustrated in Figure~\ref{fig:Deg_distribution}.

\subsubsection{Graphlets}

The final preprocessing step involves extracting graphlets, which are small, connected, non-isomorphic subgraphs that capture local structural patterns within the KG. Rather than performing a simple random walk, the KG is partitioned into these graphlets to serve as the basis for QA generation. The process considers all 29 unique graphlet structures containing 3–5 nodes (as 2-node structures are trivial), illustrated in Figure~\ref{fig:graphlets}. Subgraph enumeration is generally computationally expensive \cite{ribeiro2021survey}, making the previously described graph reduction techniques advantageous. To efficiently identify graphlets, the \texttt{graph-tool} library \cite{peixoto_graph-tool_2014} is used. The pipeline first loads the KG as an undirected graph and removes parallel edges. Next, the frequency of each graphlet shape is counted using \texttt{gt.motifs()}, and the sampling strategy of Wernicke \cite{4015377} is applied to target 10,000 occurrences per graphlet. This controls the dataset size, which is critical given that the most frequent graphlet appears over 1.8 trillion times. This process results in a final dataset of 269,574 graphlets, serving as the foundation for generating the initial QA dataset.

\subsection{Dataset Generation} \label{sec:ds_gen}

With the graphlets prepared, the central stage of the framework begins: the large-scale generation of KGQA pairs. The core of this stage is a multi-module prompt structured around knowledge-injecting graphlets, which guides and controls the LLM's generation process. In this approach, each graphlet inherently encodes the question and answer; the question targets one or two \textit{Question Nodes}, while \textit{Hidden Nodes} facilitate the reasoning required to infer the \textit{Answer Node}. The prompt explicitly includes the graphlet structure and node names to provide a factual anchor. The edge type is omitted, as it was found that allowing the LLM to infer the complex relationship yields better results than providing a simple predicate like \texttt{has\_effect}.

\begin{figure*}[htbp]
    \centering
    \includegraphics[width=\textwidth]{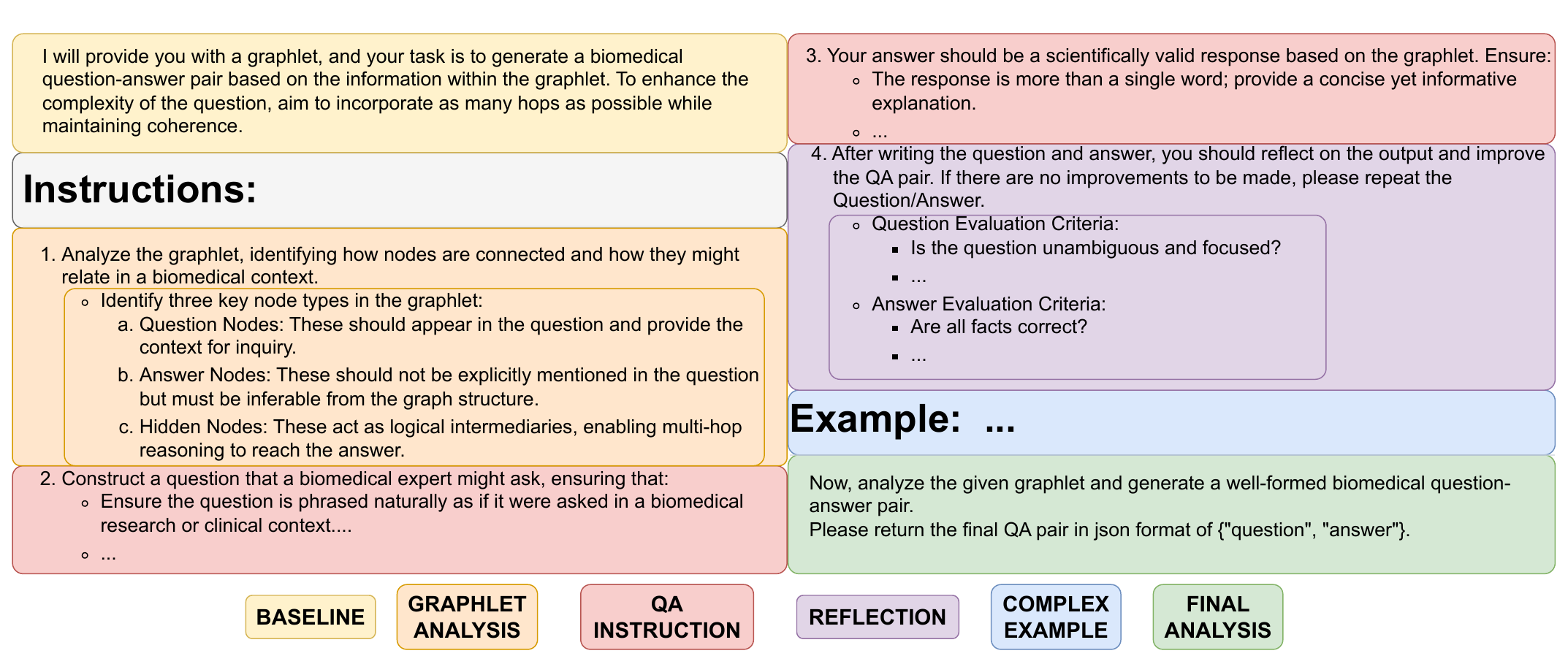}

    \caption{Compressed version of the selected prompt, showcasing extracts from some of the different modules. The \href{https://github.com/ieeta-pt/BioGraphletQA/blob/master/_prompts/generation_prompt.md}{full prompt} and the \href{https://github.com/ieeta-pt/BioGraphletQA/blob/master/_prompts/promp_ablation_modules.json}{modules} can be inspected in our GitHub Repository.}
    \label{fig:prompt}
\end{figure*}

To select and validate the prompt for dataset generation, a modular prompt building approach was employed, using an LLM-as-a-judge to systematically score outputs from 15 distinct, modular prompt configurations against six quality criteria: (1) the answer is not present in the question, (2) the question avoids graphlet terminology, (3) the answer avoids graphlet terminology, (4) the question is scientifically accurate, (5) the answer is scientifically accurate, and (6) the answer properly addresses the question\footnote{Detailed evaluation criteria is available in the \href{https://github.com/ieeta-pt/BioGraphletQA/blob/master/_prompts/promp_ablation_evaluation.md}{code repository}.}. This evaluation was conducted across 1,000 graphlets, incorporating modules such as guided reasoning instructions inspired by Chain-of-Thought and self-reflection where the model critiques its own output. This process revealed that a comprehensive full prompt, which integrated all tested modules, was the best-performing configuration. This prompt is depicted in Figure~\ref{fig:prompt}.

% }

With the graphlets prepared and the optimal prompt identified, large-scale data generation was performed using a 4-bit quantized version of the Llama-Nemotron-70B model\footnote{\href{https://huggingface.co/ibnzterrell/Nvidia-Llama-3.1-Nemotron-70B-Instruct-HF-AWQ-INT4}{Nvidia-Llama-3.1 model on Hugging Face}}. This process initially yielded 269,574 raw question-answer pairs, which were then subjected to a multi-stage refinement protocol to ensure quality and consistency. The initial filtering step addressed structural integrity, where 543 outputs that failed to parse as valid JSON were discarded. Subsequently, outliers in the length of the generated text were addressed through a statistical filtering method based on Z-scores. This step was designed to remove exceptionally short or long entries that could represent low-quality or anomalous data points. Specifically, any QA pair where either the question or the answer length fell more than three standard deviations from the mean was culled. This criterion established acceptable length ranges as 79-365 characters for questions and 59-997 characters for answers, leading to the exclusion of an additional 4,658 pairs.

\subsection{Post-Generation Filtering} \label{sec:filtering}

To improve dataset quality, LLM-based filtering was applied after the initial generation phase. Although the prompt used for generation included a reflection phase, several issues could still arise. For example, some graphlets may not contain a valid QA pair worth generating, or the generated answer may be incomplete, requiring additional knowledge to be fully correct. To mitigate these and other potential issues, automatic filtering was applied to the dataset.

\begin{figure*}[htbp]
    \centering
    \includegraphics[width=\textwidth]{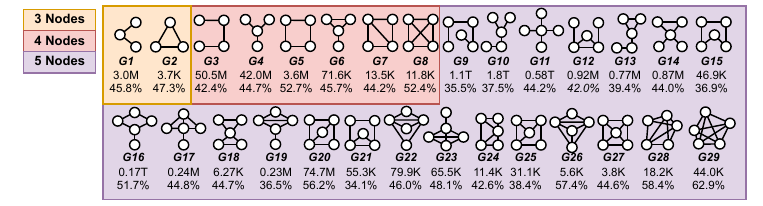}
    
    \caption{The 29 graphlet shapes with 3-5 nodes. Each shape is shown with the number of graphlets initially present (sampled to 10,000) and the acceptance ratio (QA pairs accepted / QA pairs generated). }
    \label{fig:graphlets}
\end{figure*}

The filtering prompt first instructs the model to evaluate the connections between the entities in the question to determine whether the question is coherent. Next, it attempts to answer the question and compares its response with the previously generated answer. This evaluation is structured in a JSON format to ensure that two boolean variables, \texttt{valid\_question} and \texttt{original\_answer\_valid}, are generated based on the model’s reasoning\footnote{The filtering prompt is available in the \href{https://github.com/ieeta-pt/BioGraphletQA/blob/master/_prompts/filtering_prompt.md}{code repository.}}. While the goal is to use KGs for grounding and reducing hallucination, this step assesses whether the QA pair remains valid based on the LLM’s general biomedical knowledge, independent of the specific graph context. After filtering, a final dataset of 119,856 QA pairs remained (45\% of the dataset after post-processing). Additionally, 17,101 outputs (6.47\%) were unparseable as JSON. The distribution of the accepted graphlets can be seen in Figure~\ref{fig:graphlets}.

\subsection{Supporting Document Retrieval}
\label{sec:retrieval}

While BioGraphletQA is grounded in structured knowledge, many biomedical QA scenarios also rely on unstructured text. To broaden our framework's applicability and enrich the final resource, we include a stage for augmenting the QA pairs with supporting documents from PubMed \cite{10.1093/nar/gkaa892}. This serves two purposes: it facilitates the evaluation of retrieval-augmented QA systems and provides verifiable textual evidence to validate the generated pairs. 

The process uses the PubMed abstract collection as its source. For each QA pair, a query is formed by concatenating the question and answer, which is then used with BM25 \cite{robertson2009probabilistic} to retrieve the top 10 relevant abstracts. Subsequently, the Qwen3-32B model \cite{yang2025qwen3technicalreport} is applied to each abstract to perform two tasks: 1) classify its relevance to the QA pair, and 2) extract the most pertinent supporting text snippets\footnote{The retrieval prompt is available in the \href{https://github.com/ieeta-pt/BioGraphletQA/blob/master/_prompts/retrieval_prompt.md}{code repository}.}. This process was highly effective, with a substantial majority of QA pairs being grounded in multiple documents; over 79\% of pairs were linked to at least two supporting documents, and over half (52.7\%) were linked to five or more. A small fraction of QA pairs (5.25\%) had no relevant documents, potentially indicating novel relationships inferred from the KG's structure that are not yet well-documented in abstracts, or model hallucinations.

\subsection{Task-Specific Rephrasing}
\label{sec:rephrasing}

The final stage of the framework extends the utility and adaptability of the generated data by aligning it with existing QA benchmarks. This is achieved by performing task-specific rephrasing on a subset of QA pairs using Qwen3-32B.\footnote{The rephrasing prompts for \href{https://github.com/ieeta-pt/BioGraphletQA/blob/master/_prompts/rephrase_PubMedQA.md}{PubMedQA} and \href{https://github.com/ieeta-pt/BioGraphletQA/blob/master/_prompts/rephrase_medqa.md}{MedQA} can be inspected in our code repository.}. For each pair, a prompt is constructed containing the original QA content, its supporting snippets, and relevant examples from the target dataset retrieved via BM25. The model then generates a new QA pair in the target format (e.g., multiple-choice or yes/no). This alignment enables robust, standardized evaluation of the impact of the resource on downstream tasks.

\section{Experimental Setup}
\label{sec:experiments}

To validate our proposed framework, we designed a series of experiments to answer two primary research questions:

\begin{itemize}
    \item \textbf{RQ1:} Can our framework produce QA pairs that are scientifically valid and complex, as judged by a domain expert?
    \item \textbf{RQ2:} Does data generated by our framework provide a tangible benefit when used to augment training data for downstream QA tasks?
\end{itemize}

\subsection{Evaluation of Data Quality (RQ1)}

To answer RQ1, we conducted a manual evaluation of data quality. The annotation was performed by a domain expert (co-author) to assess key characteristics. We created a validation set of 116 QA pairs using stratified sampling: for each of the 29 graphlet shapes, we randomly selected three pairs that our automated filter had \textbf{accepted} and one pair it had \textbf{rejected}. The annotator, blind to the quality filtering, rated each pair on a 5-point Likert scale across several criteria, including Scientific Validity, Question Complexity, and Answer Completeness\footnote{The full evaluation criteria is in the \href{https://github.com/ieeta-pt/BioGraphletQA/blob/master/_prompts/human_evaluation_criteria.md}{code repository}.}. The annotator had the choice to not evaluate a pair if it did not fall in their domain of expertise, which occurred 10 times, leaving 106 annotated pairs.

\subsection{Evaluation of Downstream Utility (RQ2)}

To answer RQ2, we evaluated the impact of using BioGraphletQA as a data augmentation resource for two established biomedical QA benchmarks: \textbf{PubMedQA} \cite{jin-etal-2019-pubmedqa}, a yes/no question answering task based on PubMed abstracts and \textbf{MedQA} \cite{jin2020diseasedoespatienthave}, a challenging multiple-choice QA task derived from medical board exams.

For our experiments, we fine-tuned BioLinkBERT-large \cite{yasunaga2022linkbert}, a strong baseline for biomedical NLP tasks. We opted for a BERT-based model over a generative LLM to ensure a fair evaluation free from potential training data contamination \cite{xu2024benchmarkdatacontaminationlarge}. Our experimental protocol involved training the model on increasing portions of the original training sets of both benchmarks and comparing the performance against models trained on the same data augmented with varying amounts (1k, 10k, and 20k samples) of our rephrased BioGraphletQA data. All experiments were run with 5 different random seeds to determine statistical significance.

\section{Results and Analysis}
\label{sec:results}

In this section, we present the results of our experiments, structured to directly answer the research questions posed in Section~\ref{sec:experiments}. We first analyze the quality of the data generated by our framework through expert human evaluation (RQ1), and then demonstrate its practical utility in downstream tasks (RQ2).

\subsection{Data Quality Analysis (RQ1)}
\label{sec:results_human}

To answer our first research question regarding the quality of the generated data, we conducted human evaluation as detailed in our experimental setup. The results strongly validate both the high quality of the final dataset and the efficacy of our automated filtering pipeline.

\begin{figure}[htbp]
    \centering
    \includegraphics[width=0.9\linewidth]{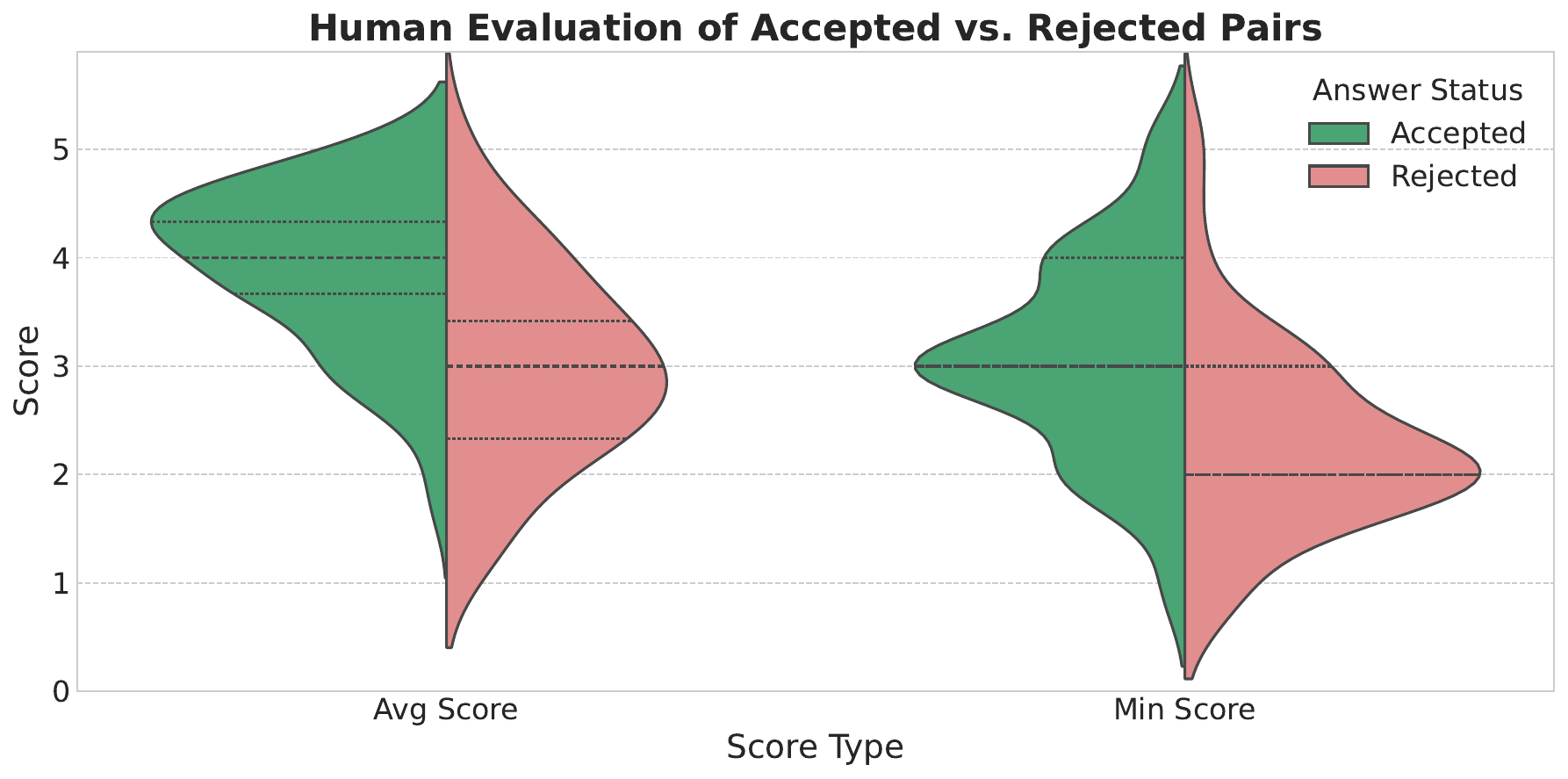}
    \caption{Human evaluation scores validating the LLM's annotation filtering. The split violin plot contrasts human-rated Average and Minimum scores for LLM annotations classified as `Accepted' (green) vs. `Rejected' (red).}
    \label{fig:violin_plot}
\end{figure}

Figure~\ref{fig:violin_plot} shows a clear distributional shift in scores between the QA pairs our LLM filter accepted versus those it rejected. The accepted pairs consistently received high scores from the human expert, while rejected pairs scored poorly. This disparity provides strong empirical evidence that Stage 3 of our framework is effective at identifying and removing low-quality data, a critical step for ensuring the reliability of synthetically generated resources.

\begin{figure}[htbp]
    \centering
    \includegraphics[width=0.85\linewidth]{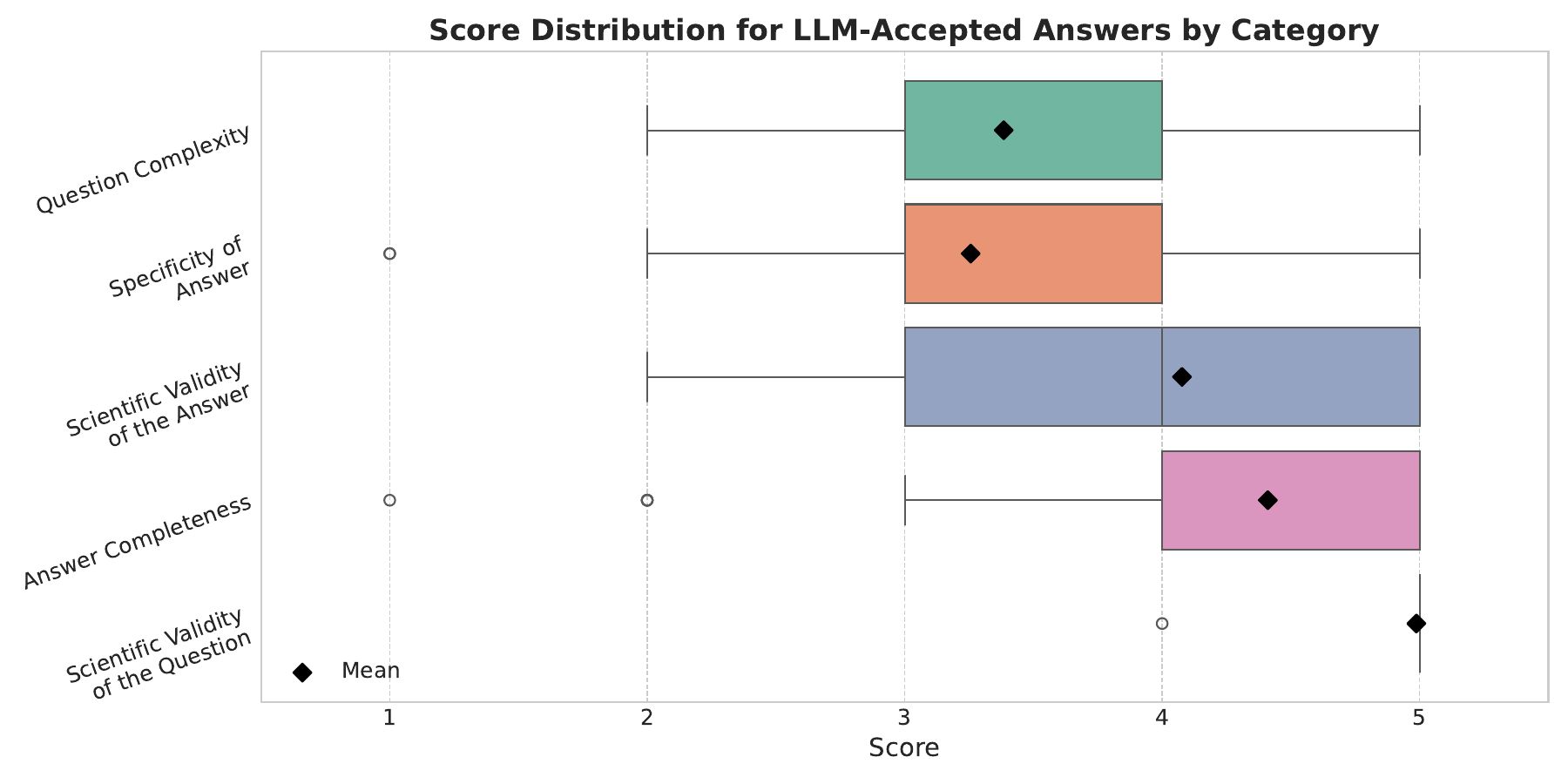}
    \caption{Boxplot of Likert based human evaluation scores across 5 categories for the accepted QA pairs.}
    \label{fig:Score_Distribution_category}
\end{figure}

Furthermore, an in-depth analysis of the accepted pairs, shown in Figure~\ref{fig:Score_Distribution_category}, confirms their high quality. All questions were rated as scientifically valid, and 88.46\% were deemed complex (score $\geq 3$). The answers also scored highly on scientific validity (92.31\% with a score $\geq 3$) and completeness (93.59\% with a score $\geq 3$). The primary area for improvement was answer specificity, where only 79.49\% of answers achieved an ideal score of 3. Despite this, 75.64\% of QA pairs achieved a minimum score of 3 across all answer criteria, confirming their overall acceptability. We consider the minimum score a particularly stringent metric, as a failure in any single category should render a QA pair invalid.

\subsection{Downstream Task Performance (RQ2)}
\label{sec:results_auto}

\begin{figure*}[htbp]
    \centering
    \includegraphics[width=\textwidth]{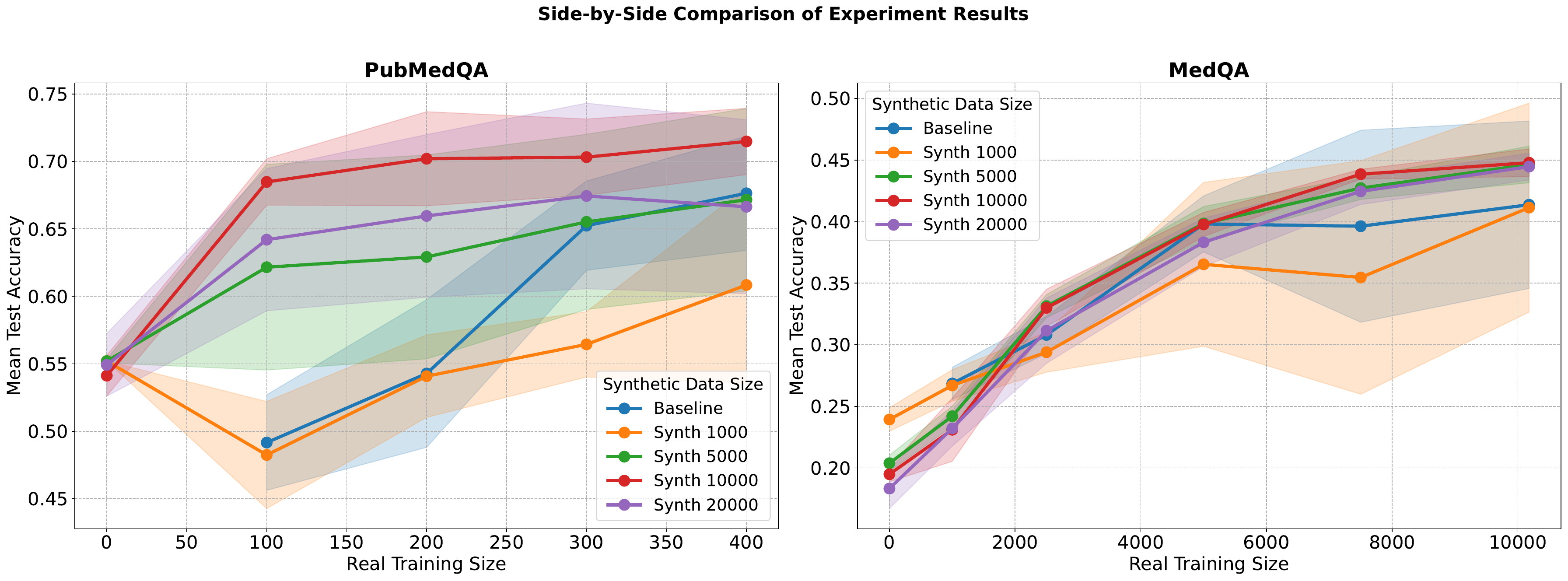}
    \caption{Effect of synthetic data augmentation on PubMedQA (left) and MedQA (right). Plots show mean test accuracy over 5 different seeds vs. real training data size. Augmenting with 10,000 synthetic samples consistently provides the largest performance gains over the baseline. Shaded areas represent standard deviation.}
    \label{fig:results}
\end{figure*}

To answer our second research question regarding the utility of our generated data, we evaluated its impact as an augmentation resource for downstream tasks. The results, presented in Figure~\ref{fig:results}, demonstrate that data generated by our framework provides benefits when used for augmenting gold data.

The findings show performance gains in both low-resource and full-resource settings. The most striking result was on PubMedQA in a low-resource setting: augmenting just 100 real samples with 10,000 synthetic pairs increased mean accuracy from 49.2\% to 68.5\%, effectively compensating for the scarcity of labeled data. This positive trend extended to the MedQA benchmark, where adding 10,000 synthetic samples to the full 10,000-sample training set improved accuracy from a 41.4\% baseline to 44.8\%. Our analysis also indicates an optimal augmentation size. An addition of 10,000 synthetic samples consistently provided the most significant and stable performance gains across both datasets. In contrast, a smaller augmentation of 1,000 samples was often unstable, while an excessive 20,000 samples showed diminishing returns and even degraded performance on PubMedQA, suggesting a saturation point for this training configuration.

\section{Discussion and Limitations}
The results of our human and automatic evaluations validate our proposed framework as an effective method for generating high-quality, complex QA data. The success of the downstream augmentation task, particularly in the low-resource PubMedQA setting, underscores the potential of knowledge-anchored synthetic data to mitigate data scarcity in specialized domains. The rich modular prompt, combined with automated filtering, offers a scalable and reliable alternative to costly manual annotation.

We acknowledge the limitation of relying on a single human annotator for our qualitative assessment; future work should involve multiple annotators to establish inter-rater reliability. Additionally, our framework could be refined to improve answer specificity, which our analysis identified as the primary area for improvement. Finally, the quality of the generated data is inherently linked to the capabilities of the underlying language model; with better computational resources, and as more advanced models become available, our framework can be used to produce datasets of even higher quality.

\section{Conclusion}

In this paper, we describe and validate a new graphlet-anchored framework for generating complex, factually-grounded question-answering data. We introduced its first instantiation, BioGraphletQA, a new, large-scale resource of 119,856 QA pairs for the biomedical domain.

Our comprehensive evaluation demonstrated the framework's success. A qualitative assessment by a domain expert confirmed the high scientific validity and complexity of the dataset and validated our automated filtering pipeline. Furthermore, quantitative experiments showed its utility as a data augmentation tool, with performance on PubMedQA increasing from 49.2\% to 68.5\% in a low-resource setting. These findings underscore the value of our framework for creating critical resources to advance complex QA tasks. We release both the BioGraphletQA dataset and our generation framework publicly to facilitate future research into reliable, factually-grounded question answering.

\begin{credits}
\subsubsection{\ackname} This work was funded by FEDER - Fundo Europeu de Desenvolvimento Regional funds through the project CENTRO2030-FEDER-02595400, and by the Foundation for Science and Technology (FCT) through contract \url{https://doi.org/10.54499/UID/00127/2025}. Richard A. A. Jonker is funded by the FCT doctoral grant PRT/BD/154792/2023, 
with DOI identifier \url{https://doi.org/10.54499/PRT/BD/154792/2023}.

\subsubsection{\discintname}
The authors have no competing interests to declare that are relevant to the content of this article
\end{credits}

\bibliographystyle{splncs04}
\bibliography{refs}

\end{document}